\documentclass[sigconf]{acmart}
\usepackage{epstopdf}
\usepackage{amssymb}
\usepackage{fixltx2e}
\usepackage{subcaption}
\usepackage{amsmath}
\usepackage{algorithm}
\usepackage{algorithmic}
\usepackage{lipsum}

\usepackage{multirow}

\def\BibTeX{{\rm B\kern-.05em{\sc i\kern-.025em b}\kern-.08emT\kern-.1667em\lower.7ex\hbox{E}\kern-.125emX}}

\copyrightyear{2018}
\acmYear{2018}
\setcopyright{acmlicensed}
\acmConference[ICCAD '19]{ICCAD '19: International Conference On Computer Aided Design}{November 4-7, 2019}{Westminster, CO}
\acmBooktitle{ICCAD '19: International Conference On Computer Aided Design, November 04--07, 2019, Westminster, CO}
\acmPrice{15.00}
\acmDOI{10.1145/1122445.1122456}
\acmISBN{978-1-4503-9999-9/18/06}

\begin{document}

%
% The "title" command has an optional parameter, allowing the author to define a "short title" to be used in page headers.
\title{On Neural Architecture Search for Resource-Constrained Hardware Platforms}

%
% The "author" command and its associated commands are used to define the authors and their affiliations.
% Of note is the shared affiliation of the first two authors, and the "authornote" and "authornotemark" commands
% used to denote shared contribution to the research.
\author{Qing Lu}
\affiliation{%
  \institution{University of Notre Dame}
}
\email{qlu2@nd.edu}

\author{Weiwen Jiang}
\affiliation{%
  \institution{University of Notre Dame}
}
\email{wjiang2@nd.edu}

\author{Xiaowei Xu}
\affiliation{%
  \institution{University of Notre Dame}
}
\email{xxu8@nd.edu}

\author{Yiyu Shi}
\affiliation{%
  \institution{University of Notre Dame}
}
\email{yshi4@nd.edu}

\author{Jingtong Hu}
\affiliation{
  \institution{University of Pittsburgh}
}
\email{jthu@pitt.edu}

%
% By default, the full list of authors will be used in the page headers. Often, this list is too long, and will overlap
% other information printed in the page headers. This command allows the author to define a more concise list
% of authors' names for this purpose.
\renewcommand{\shortauthors}{Qing Lu and Weiwen Jiang, et al.}

%
% The abstract is a short summary of the work to be presented in the article.
\begin{abstract}
In the recent past, the success of Neural Architecture Search (NAS) has enabled researchers to broadly explore the design space using learning-based methods. Apart from finding better neural network architectures, the idea of automation has also inspired to improve their implementations on hardware. While some practices of hardware machine-learning automation have achieved remarkable performance, the traditional design concept is still followed: a network architecture is first structured with excellent test accuracy, and then compressed and optimized to fit into a target platform. Such a design flow will easily lead to inferior local-optimal solutions.
To address this problem, we propose a new framework to jointly explore the space of neural architecture, hardware implementation, and quantization.
Our objective is to find a quantized architecture with the highest accuracy that is implementable on given hardware specifications. We employ FPGAs to implement and test our designs with limited loop-up tables (LUTs) and required throughput. Compared to the separate design/searching methods, our framework has demonstrated much better performance under strict specifications and generated designs of higher accuracy by 18\% to 68\% in the task of classifying CIFAR10 images. With 30,000 LUTs, a light-weight design is found to achieve 82.98\% accuracy and 1293 images/second throughput, compared to which, under the same constraints, the traditional method even fails to find a valid solution.
\end{abstract}

\maketitle

\setlength{\textfloatsep}{6pt}
\setlength{\floatsep}{6pt}
\setlength{\dbltextfloatsep}{6pt}

\section{Introduction}
% The success of machine learning in a variety of applications has not only demonstrated its strong ability \cite{xu2018scaling, xu2018efficient, tianchen2019, li2019exploiting}, but also led to the ever-growing demand in the off-the-shelf solutions to specific machine learning systems \cite{xu2019whole,wang2019msu}. In consequence, a trend from hand-crafted to automatic machine learning is observed in the very recent years. Particularly, the design of deep neural networks, whose performance heavily depends on their architectures, can now be designed by an automatic framework named neural architecture search \cite{zoph2016neural}. Without any human intervention, the NAS network has exhibited parallel performance with the best design by experts.

Machine learning has demonstrated great success in a variety of applications \cite{xu2018scaling, xu2018efficient, tianchen2019, li2019exploiting}, which leads to the ever-growing demand in the off-the-shelf solutions to application-specific systems \cite{xu2019whole,wang2019msu,jiang2018heterogeneous,jiang2019xfer}.
Designing neural networks applying the hand-crafted approach, however, involves huge expertise and labor.
In response to this challenge, automated machine learning (Auto-ML) is proposed to build neural networks without human intervention; in particular, Neural Architecture Search (NAS) \cite{zoph2016neural} is proposed to identify the neural architecture with competitive or even better accuracy against the best design explored by experts.

On the other side, when deploying architectures explored by NAS to real-world platforms, such as AIoT \cite{aiot} and mobile embedded platforms \cite{jiang2016optimal,jiang2017optimal,tan2018mnasnet,cai2018proxylessnas,wu2018FBNet}, it is inevitably limited by the hardware constraints. As a result, hardware-aware machine learning \cite{jiang2019accuracy,cai2018proxylessnas,wu2018FBNet,ZhangJSH19} has emerged to explore neural architectures with the consideration of hardware efficiency on a target fixed hardware design.
Most recently, authors in \cite{jiang2019hardware} open the hardware space in NAS to jointly explore the architectures and hardware designs.
However, almost all existing methods adopt a separated optimization flow \cite{koeplinger2016automatic}: a large network is first invented with excellent performance, and then compressed and optimized to fit into a target platform.
Note that compression techniques, especially quantization \cite{xu2018efficient,xu2017edge, xu2018resource,xu2018quantization}, have to be considered to fit the model into resource-constrained hardware platforms, which can tremendously reduce the hardware resource consumption and related computation consumption.
Consequently, such approach usually failed to find the overall optimal solutions. For example, the best quantization scheme specifically tuned for a network may be significantly inferior when applied to another network or even not implementable under certain hardware specifications.

In this paper, we delve into the NAS-based methods of design automation on hardware-constrained platforms. We aim to answer such a concrete question: for a specific task, what is the best neural architecture with the highest accuracy that is implementable given a defined set of hardware specifications? In particular, a novel co-exploration framework is proposed to investigate the optimality of neural architectures with quantization. In our framework, we parameterize the layer-wise quantization and search these parameters jointly with the hyperparameters of the architecture. A hardware model is built by searching the hardware space and validated by the design specifications. We use FPGAs as the target platform and run experiments under various configurations and specifications. Compared with the existing separately searching, the proposed joint search method is more robust, achieving 18\% to 68\% higher accuracy on common used data sets.

% It turns out the experimental result identifies the weakness of separately searching or designing the architecture and implementation, under different hardware specifications, where the joint search method are observed to be more robust.

The remainder of this paper is organized as follows. In Section \ref{sec:today}, we outline the progress in neural architecture search associated with hardware design. After that, we present the details of our design framework in Section \ref{sec:framework}. Section \ref{sec:exp} then investigates the performance of our framework by experiment as compared with conventional methods of separate search. Finally, Section \ref{sec:con} remarks the conculsion and future work.

\begin{figure}[t]
  \centering
  \includegraphics[width=2.6 in]{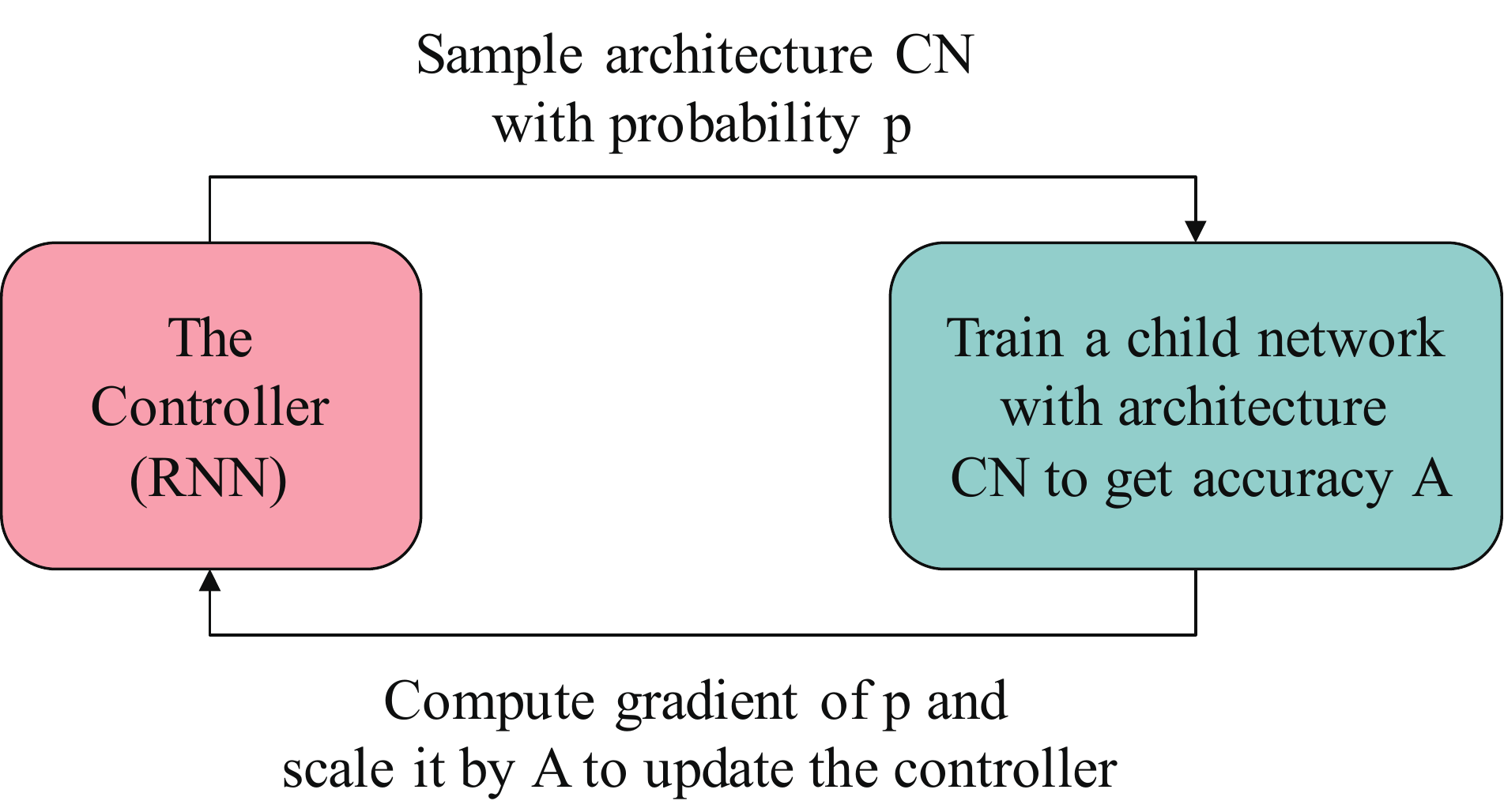}
  \vspace{-10pt}
  \caption{The Pure-Software NAS framework.}\label{Fig:NASOriFramework}
\end{figure}
\section{Today's NAS: from Pure-Software via Hardware-Aware to Co-Design}\label{sec:today}

Likewise the development of designing the embedded systems, the evolution of today's NAS has gone through three phases: (1) exploring structure only, called pure-software NAS in this paper; (2) considering efficiency on a fixed hardware in exploring structures, called hardware-aware NAS; (3) co-exploring hardware implementation and structures, called Co-Design NAS.
In the following, we will introduce each phase in detail, and then outline the development trend of NAS in the future.

\textbf{Pure-Software NAS.} Figure~\ref{Fig:NASOriFramework} shows the NAS framework presented in \cite{zoph2016neural}.
In NAS, a controller (implemented as an RNN) iteratively generates a child network and obtains its accuracy $A$ by training it on a held-out data set.
Then, accuracy $A$ will be used as the reward signal to the controller for its self-evolution for the next iteration.
The search process will be stopped if the controller is converged for the maximum accuracy, or a termination condition is satisfied. Existing work has demonstrated that the automatically generated network architectures can achieve close accuracy to the best human-invented architectures on the image classification task \cite{zoph2018learning,zoph2016neural}.

\emph{Search Space:} For image classification, the linear array is applied as the backbone of network architecture.
In \cite{zoph2016neural}, each cell is a normal convolution operation.
In each cell, the search space is composed of the filter size, strides, and the number of filters.
In \cite{zoph2018learning}, authors propose to incorporate $B$ blocks ($B=5$ in the paper) in one cell, where each block is a 2-branch structure, mapping from 2 input tensors to 1 output tensor.
And the controller determines the type of operation on each input tensor.
The operations include the different size of depthwise-separable convolution, atrous convolutions, average pooling, max pooling, skip connection, etc.

\begin{figure}[t]
  \centering
  \includegraphics[width=3.3 in]{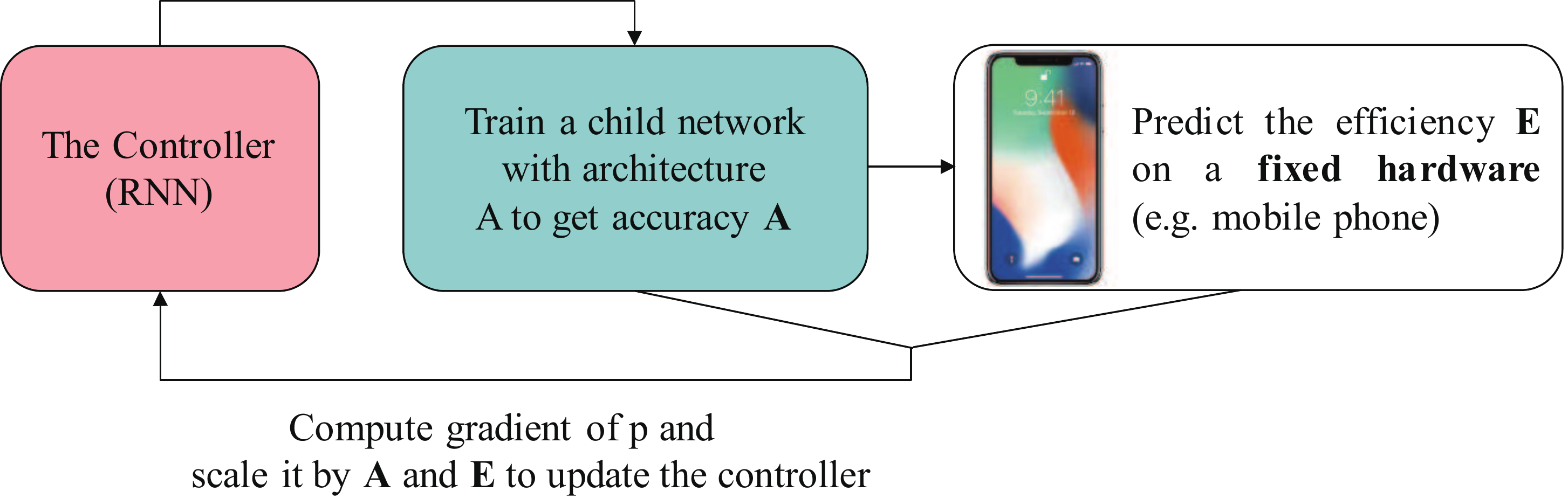}
  \vspace{-10pt}
  \caption{The Hardware-Aware NAS framework.}\label{Fig:HW-NAS-P1}
\end{figure}

\textbf{Hardware-Aware NAS.} Figure \ref{Fig:HW-NAS-P1} illustrates the works on searching neural architectures targeting for fixed hardware \cite{tan2018mnasnet,cai2018proxylessnas,wu2018FBNet}.
In these works, mobile phones are commonly be employed to be the testbed.
In order to guarantee the final system can satisfy the timing specification.
The framework will test the hardware efficiency (e.g., latency, energy consumption) for each child network.
As shown in Figure \ref{Fig:HW-NAS-P1}, after training, the child network will be sent to the target platform to be executed.
During execution, the hardware efficiency $E$ will be profiled.
$E$ together with the accuracy $A$ will be applied to update the controller to explore a better neural network architecture.

More specifically, authors in \cite{tan2018mnasnet} propose two optimization methods.
Assume the hardware efficiency $E$ stands for latency.
Given the latency specification $S$, the first method is to maximize the accuracy $A$, subjecting to the constraint of $E\le S$.
With this method, it still has the mono-criteria on maximizing accuracy.
This method can guarantee the hardware efficiency to meet the specifications, but it cannot provide the Pareto optimal solutions.
Then, a weighted product method to approximate Pareto optimal solutions is proposed.
The objective function is revised as $max=A\times \frac{E}{S}^w$, where $w$ is the weight factor.
In this way, it enables the controller to effectively approximate Pareto solutions nearby the specification $S$.

All the above approaches consider the hardware efficiency during the search space. However, they neglect the hardware design freedom, which is commonly given in many AI applications (e.g., IoT, embedded systems).
As a result, it will potentially lead to inferior solutions.
A more elegant way to tailor the hardware design for neural architecture needs to be exploited.

\textbf{Co-Design NAS.} Most recently, we propose the hardware/software co-design NAS to simultaneously optimize architecture accuracy and hardware efficiency.
Interestingly, we observe that the hardware design space is tightly coupled with the architecture search space, i.e., the best neural architecture depends on the hardware (hardware-aware NAS), and the best hardware depends on the neural architecture. It is therefore best to jointly explore both spaces to push forward the Pareto frontier between hardware efficiency and test accuracy for better design tradeoffs.

\begin{figure}[t]
%\vskip 0.2in
\begin{center}
\centerline{\includegraphics[width=\columnwidth]{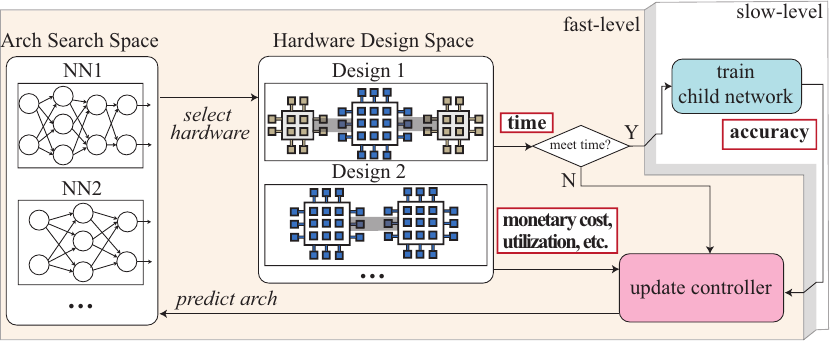}}
\vspace{-10pt}
\caption{The Co-Design NAS framework.}\label{Fig:IntrIll}
\end{center}
\end{figure}

Specifically, our architecture search space and hardware design space co-exploration framework is shown in Figure~\ref{Fig:IntrIll}(b).
The proposed co-exploration can be built on any existing NAS framework \cite{zoph2016neural,cai2018proxylessnas,liu2018darts,bender2018understanding} by expanding it to delve into the hardware design space, where a two-level (fast and slow) exploration is iteratively conducted.
In the fast exploration, the best hardware design is identified for the sampled neural architectures without lengthy training.
The architectures with inferior hardware efficiency will be quickly pruned, which significantly accelerates the search process.
Thereafter, the superior candidates are trained in the slow exploration for controller update using policy gradient reinforcement learning to explore the coupled architecture search space. The optimization objectives in the hardware design space can be varied according to the design specifications, such as area, monetary cost, energy efficiency, reliability, resource utilization, etc.

\textbf{Near Future.} The essential objective of NAS is for AI democratization.
Although the Co-Design NAS has already significantly pushed forward the progress to automatically implement machine learning tasks on hardware, without considering the constrained hardware resource in edge computing where tons of AI applications are waiting to be deployed, it will easily find inferior solutions or even cannot find feasible solutions.
In the following sections, we made innovations on NAS to propose quantization search for resource-constrained hardware on edge.

\section{Tomorrow's NAS: Landing on Edge with Quantization Search for Resource-Constrained Hardware}\label{sec:framework}

% \section{Neural Architecture and Quantization Search}\label{sec:space}
This section will present our framework on NAS and hardware co-design.
Specifically, we target on jointly optimizing neural architectures together with their quantization and hardware designs with multiple objectives, which can guarantee the resultant implementation to meet the given specifications.

The overall framework is illustrated in Figure \ref{fig:overview}. We use the controller to explore the architecture space and the hardware search tool to explore the hardware space. In each episode, the controller samples a child network architecture as well as its quantization scheme. Based on this network, the hardware builder will perform a search procedure through the hardware space for the model of an FPGA-based design. During this process, each candidate model is validated by the design specifications and accordingly the result is used to generate the return to the controller. If any FPGA model is valid, the sampled quantized network is trained on a held-out dataset and feedback the controller with its test accuracy, and otherwise, the return is zero instead. The following sections will reveal the details of this framework.
% investigating joining the quantization search into the architecture search process and optimizing the quantized neural architecture in accuracy.
% In addition, we consider multi-objective hardware design in our framework to guarantee that the effectiveness of the resultant implementation can satisfy the given specifications.

\subsection{Design Space and Parameterization}
This paper takes the widely used convolutional neural networks and their FPGA implementation to show the proposed framework, where a serial of stacked convolutional layers are optimized and implemented on an FPGA.
The proposed framework will jointly consider three design spaces: architecture space, quantization space and hardware space.

% \todo{Design space size}
%  $\mathcal{A}$  $\mathcal{Q}$  $\mathcal{H}$
% Since the layers are stacked, the architecture and quantization space can be simply based on the parameterization of each single layer and expanded by their repetition with their sizes rising exponentially. On the other hand, however, the hardware space is dependent on the specific selections from $\mathcal{A}$ and $\mathcal{Q}$, and needs to be explored as a whole. Therefore, the total design space for $L$ (convolutional) layers are represented as
% \begin{equation}
%     \mathcal{S} = \mathcal{A}^L \times \mathcal{Q}^L\times \mathcal{H}(\mathcal{A}, \mathcal{Q}, L)
% \end{equation}

% A typical convoluntional neural network designed for image classification tasks has a serial structure consisting of deeply stacked convolutional layers, followed by 1-3 fully connected layers. In this paper, we focus on accelerating the convolution operations as they not only occupy the absolute majority of whole computation workload but also cast a determining impact on the overall performance of the network.

% According to our methodology,

\subsubsection{Architecture Space}
We consider one neural network layer is composed of a convolutioanl operation followed by a pooling operation.
For a convolutional operation, its exploration space can be parameterized, including the number of filters ($N$), filter height ($Fh$), filter width ($Fw$), stride height ($Sh$), and stride width ($Sw$).
For a pooling operation, we employ the size parameter $Ps$ to indicate its length and stride.
As a whole, each layer can be represented by a 6-element sequence: $(N, Fh, Fw, Sh, Sw, Ps)$, and the architectural space of each layer is
$\mathcal{A}=\prod_{p\in A}|p|,$
where $A = \{N, Fh, Fw, Sh, Sw, Ps\}$, and $|p|$ denotes the number of possible values of a parameter $p$.

\subsubsection{Quantization Space}
We apply the quantization to all the trainable parameters and activations in each layer to make tradeoffs between the hardware size and test accuracy.
In this paper, we consider a linear quantization with fixed-point representation that is composed of the integer and fractional parts which are taken as separate parameters in our framework.

Assuming the rectified linear unit (ReLU) as the activation function, the output $A$ of the convolutional layer is non-negative, and we apply the unsigned quantization as
\begin{equation}
    Q(A) = clip(round(\frac{A}{\Delta q})\times \Delta q,\ 0,\ B - \Delta q).
\end{equation}
where $\Delta q$ is the precision and $B$ is the range amplitude, both of which are determined by the bit width in the integer part $Ai$ and fractional part $Af$, respectively.
We conclude their relationship as: $B = 2^{Ai},\ \Delta q = 2^{-Af}.$

For the weight and bias parameters $W$, signed quantization is applied, such that
\begin{equation}
    Q(W) = clip(round(\frac{W}{\Delta q})\times \Delta q,\ -B,\ B - \Delta q),
\end{equation}
where we have the relationship between $\Delta q$, $B$, $Wi$ and $Wf$ as follows: $B = 2^{Wi-1},\ \Delta q = 2^{-Wf}.$

Similarly to the architecture parameterization, the quantization scheme can be represented by $Q = \{Ai, Af, Wi, Wf\}$ and thus the quantization space is $\mathcal{Q}=\prod\limits_{p\in Q}|p|.$

% We apply the quantization to all the trainable parameters and activations of each layer separately. The quantization parameters have a crucial impact on the inference accuracy and meanwhile they also scales the hardware size as they determine the data width, operation logics, memory requirement, and so on.

\subsubsection{Hardware Space}
Given a determined architecture together with its quantization, the implementation varies in terms of two aspects: intra-layer parallelism (single-layer accelerator design) and inter-layer parallelism (mapping accelerators to an FPGA).

For the single-layer accelerator design, we adopted the widely used tile-based paradigm \cite{Zhang:2015:OFA:2684746.2689060}.
We represent tiling parameters as a sequence of functions with ${Tm}(M)$ as the number of channels of the input tiles, ${Tn}(N)$ as the number of channels of the output tiles, $Tr(R)$ as the height of input tile, and $Tc(C)$ as the width of the input tile, where $M$, $R$ and $C$ are the number of channels, rows, and columns of the input feature maps, respectively.

For the mapping of single-layer accelerators to an FPGA, we partition and allocate hardware resources to accelerators.
The partition scheme $P$, as a function of $L$, is a selection from all the combinations of the $L$ layers clustered into any number of sections from 1 to $L$, which results in a space consisting of $2^{L-1}$ candidates. The hardware space is then represented by $\mathcal{H}(L) = 2^{L-1}\prod\limits_{p\in H}|p|$, where $H = \{Tm, Tn, Tc, Tr\}$.

% Because the loops over dimension $Fh$ and $Fw$ are typically small, we select them to be unrolled to improve parallel computation and hence will not parameterize them.

% , and the resultant parameters are the tile sizes in all dimensions upper-bounded by the corresponding architecture parameters actually sampled. Accordingly, we represent these parameters as a sequence of functions with ${Tm}(M)$ as the number of channels of the input tiles, ${Tn}(N)$ as the number of channels of the output tiles, $Tr(R)$ as the height of input tile, and $Tc(C)$ as the width of the input tile, where $M$, $R$ and $C$ are the number of channels, rows, and columns of the input feature maps, respectively. Because the loops over dimension $Fh$ and $Fw$ are typically small, we select them to be unrolled to improve parallel computation and hence will not parameterize them.

% Once the realization of all the convolutional layers is decided, the global optimization is still needed for mapping these single-layer accelerators onto the target FPGA, where we partition and allocate hardware resources to them according to the partition scheme. The partition scheme $P$, as a function of $L$, is then a selection from all the combinations of the $L$ layers clustered into any number of sections from 1 to $L$, which results in a space of $2^{L-1}$ candidates. The totality of hardware space is then
% $$\mathcal{H}(L) = 2^{L-1}\prod\limits_{p\in H}|p|,$$
% where $H = \{Tm, Tn, Tc, Tr\}$.

\begin{figure}[t]
    \centering
    \includegraphics[width=0.45\textwidth]{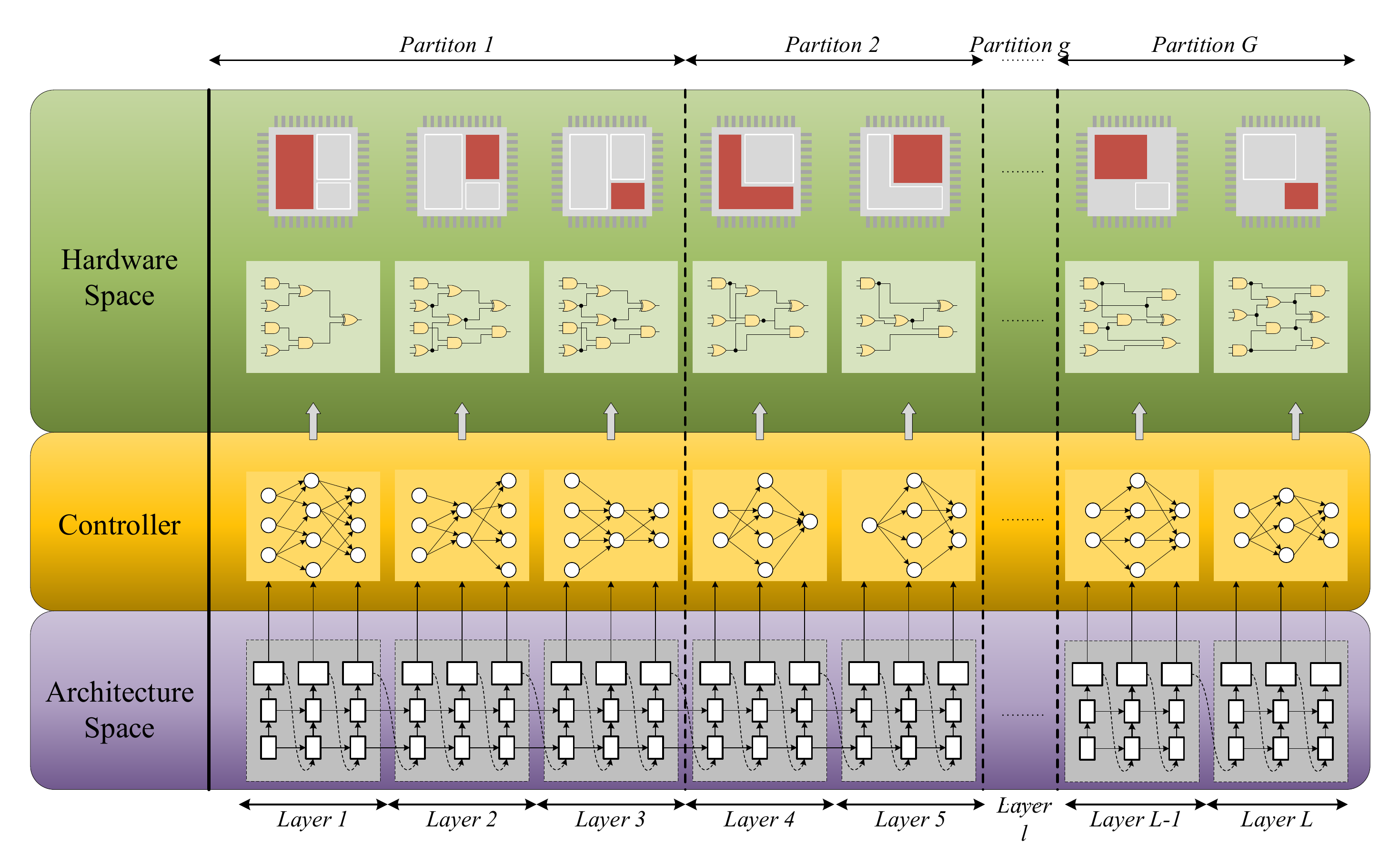}
    \vspace{-10pt}
    \caption{Overview of the proposed exploration framework.}
    % This figure illustrates the design space mapping in layers from the perspective of the controller.}
    \label{fig:mapping}
\end{figure}

\subsubsection{Overall} The proposed framework will jointly determine architecture, quantization, and tiling parameters, together with partition of layers, to identify the neural architecture and hardware implementation, such that both test accuracy and hardware efficiency can be maximized.

We will use the reinforcement learning method to explore architecture and quantization spaces, and develop a multi-objective search algorithm to explore the hardware space.
More specifically, there is a controller to control the exploration, as shown in Figure~\ref{fig:mapping}.
Details will be discussed in the following sections.

% The relationship between the spaces and the controller is shown in Figure~\ref{fig:mapping}. The details of our method will be discussed in the following sections.

% According to all of the above three design spaces, we can calculate the size of the whole design as $\mathcal{S}=2^{L-1}\prod\limits_{p\in A\cup Q\cup I}|p|.$

% the objective of our design of an $L$-layer CNN accelerator on the resource-constrained hardware is to decide the following parameters:
% \begin{itemize}
%     \item for each layer, a set of architecture parameters $a$ = ($N$, $Fh$, $Fw$, $Sh$, $Sw$, $Ps$);
%     \item for each layer, a set of quantization parameters $q$ = ($Ai$, $Af$, $Wi$, $Wf$);
%     \item for each layer and a given set of $a$, a set of hardware parameters $h$ = ($Tm$, $Tn$, $Tr$, $Tc$); and
%     \item a way of partitioning $L$ layers.
% \end{itemize}{}
% As a result, the total design space is $$\mathcal{S}=2^{L-1}\prod\limits_{p\in A\cup Q\cup I}|p|.$$ We use the NAS method to explore the architecture and quantization space for maximizing the performance in accuracy, and develop a hardware model with a multi-objective search algorithm to explore the hardware space for validating design samples. The relationship between the spaces and the controller is shown in Figure~\ref{fig:mapping}. The details of our method will be discussed in the following sections.

\subsection{Update the Controller}\label{sec:method}
\begin{figure}[t]
    \centering
    \includegraphics[width=0.45\textwidth]{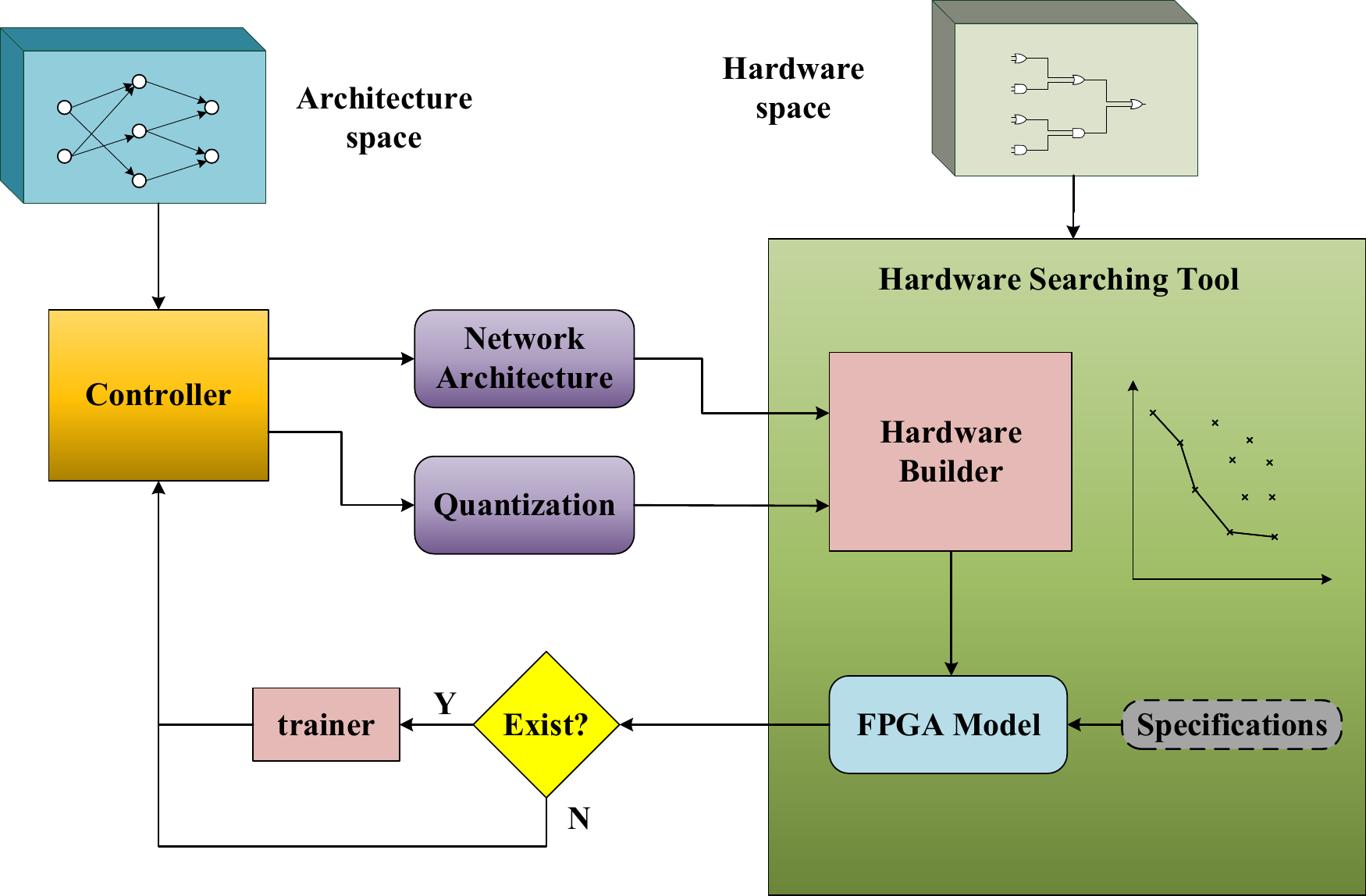}
    \vspace{-10pt}
    \caption{Hardware-architecture co-design framework.}
    \label{fig:overview}
\end{figure}

The architecture and quantization parameters are both optimized to generate high accuracy. As shown in Figure \ref{fig:overview}, we employed reinforcement learning method to explore the space $\mathcal{A}$ and $\mathcal{Q}$ where the controller interacts with the environment modeled as a Markov Decision Process (MDP). In each episode, the controller rolls out a sequence of actions under a stochastic policy. These actions, used as the architecture parameters $A$ and quantization parameters $Q$, are mapped to a \emph{quantized} child network. Next, we evaluate the sampled child network in two stages. In the first stage a hardware searching tool is developed to verify whether the sampled network is implementable under the constraints of design specifications. If the result is positive, i.e. there exists an implementable hardware model, and the second stage will launch to train and validate the child network on a held-out dataset. After the child network validation is finished, a reward signal
\begin{equation}
    R (a, q) = \left\{
    \begin{array}{cc}
        0, & \mathcal{H}(a, q) = \varnothing  \\
        Acc, & \text{otherwise}
    \end{array} \right.
\end{equation}
is returned to the controller for updating. In the above formula, $\mathcal{H}(a, q)$ represents the hardware space given sampled parameters $a$ and $q$.
We follow the Monte Carlo policy gradient algorithm \cite{WILLIAMS1992Simple} to update the controller using
\begin{equation}
    \triangledown J(\theta) = \frac{1}{m}\sum\limits_{k=1}^m\sum\limits_{t=1}^T\gamma^{T-t}\triangledown_\theta\log \pi_\theta(a_t|a_{\left(t-1\right):1})(R_k-b)
\end{equation}
where $m$ is the batch size and $T$ is the total number of steps in each trajectory. The rewards are discounted at every step by an exponential factor $\gamma$ and the baseline $b$ is the exponential moving average of the rewards.
\subsection{Co-Explore Architecture and Quantization}
\begin{figure*}
    \centering
    \includegraphics[width=\textwidth]{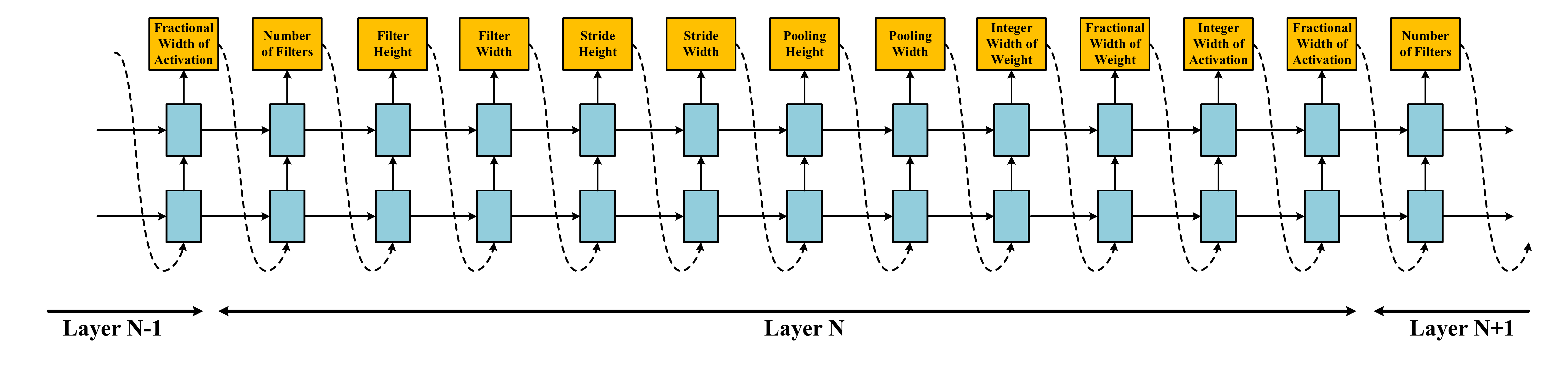}
    \vspace{-20pt}
    \caption{The controller in sync search samples both architecture and quantization parameters layer by layer.}
    \label{fig:controller}
\end{figure*}
\begin{figure}[t]
    \centering
    \includegraphics[width=0.45\textwidth]{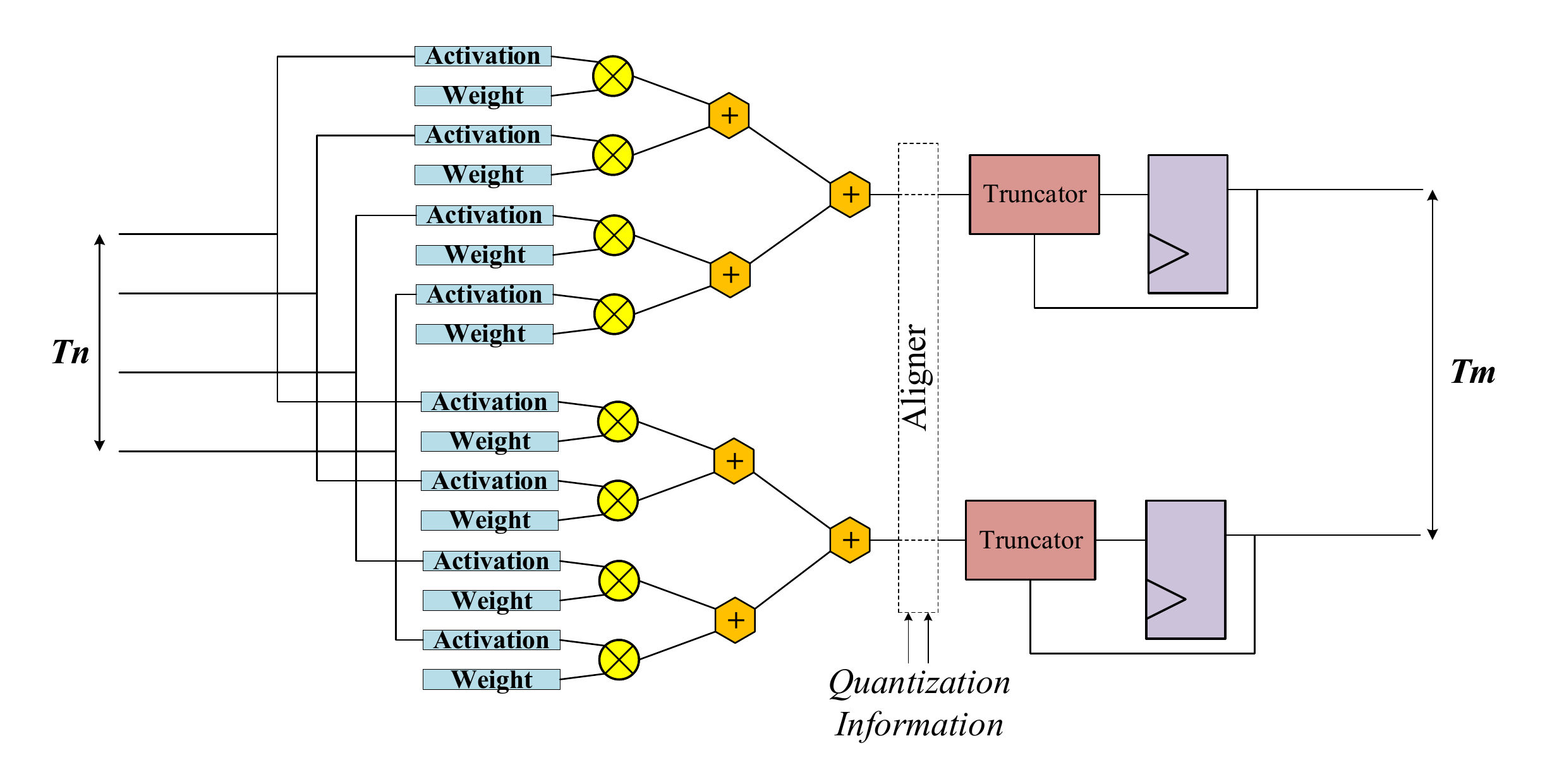}
    \vspace{-10pt}
    \caption{RTL architecture of quantized computing engine.}
    \label{fig:qce}
\end{figure}
Much like the architectures, the quantization also determines the overall performance and computational complexity of a network. Therefore, it is natural to automate the design of quantization together with the design of the architecture. Under constrained resource in hardware,
the joint exploration of architecture and quantization space is actually the optimization of the trade-off between structural complexity and data cleanness. Therefore, the reward signal is the reflection of how efficiently the hardware freedom are utilized.

The implementation of architecture-quantization joint search may vary by different settings and discretion, but generally there are two types of methods characterized by the number of controller used: a single controller to predict both architecture and quantization, or two controllers to predict them separately. In this paper, we focus on the single-controller method and extend the RNN-based controller in \cite{zoph2016neural}. As displayed in Figure~\ref{fig:controller}, we simply insert 4 additional steps into the controller, each step sampling one of the aforementioned quantization parameters.

For comparison, we list all the plausible methods for performing the automate design of neural architecture with quantization. The difference among these methods reside in the space to be explored.
\begin{itemize}
    \item[-] \emph{quantization} search. This is the traditional method that employs the controller to search only the quantization for a given architecture.
    \item[-] \emph{architecture} search. This is the reverse procedure of quantization search, where the quantization is fixed and the objective is to find the architecture to best fit this quantization.
    \item[-] \emph{joint} search. This is the exploration of architecture and quantization as related space and what we intend to investigate in this project.
\end{itemize}
\subsection{Explore Hardware Space}
We revisit the hardware space discussed in Section \ref{sec:framework}, this time with consideration of our hardware model. In our specific model, the specifications are in the LUT usage of the target FPGA and the throughtput in frames-per-second, though the framework can be adapted to other aspects without much effort. In particular, only $Tn$ and $Tm$ are variable while $Tr$ and $Tc$ are dropped due to their irrelevance to the target specifications. The other parameters such as quantizations are given as constant instead of variable. As a result, the actual hardware space to explore is
\begin{equation}\label{eq:ispace}
    \mathcal{H}(L) = 2^{L-1}\prod\limits_{i=1}^L N_{i-1}N_{i}
\end{equation}
whose size grows exponentially with $L$. To avoid exploring $\mathcal{H}$ exhaustively, we have developed an efficient searching algorithm using dynamic programming. The hardware model and searching algorithm are explained in details as follows.
\subsubsection{Tile-Based Implementation}
In the tile-based model, the main processing unit is the quantized computation engine (QCE) which is composed of an array of multipliers, an adder tree, a truncator, and the accumulation registers (Figure \ref{fig:qce}). For implementation on the FPGAs, the consumption of lookup-tables (LUTs) by each QCE scales with $Tn$, $Tm$, and the bit-width of the layer as it configures the size of the above components.
Due to the data format inconsistency between the activation and weight (inter-layer), and the inconsistency between activations of consecutive layers (intra-layer), we customize the QCEs in each layer, according to the 6 parameters $Wi$, $Wf$, $Ai$, $Af$, $Ai^{\prime}$, and $Af^{\prime}$, where $Ai^{\prime}$, and $Af^{\prime}$ are the bitwidth of the activation from the last layer. For the inter-layer inconsistency, the problem is handled by informing the multipliers and adders of the number of integer and fractional bits of each operand. This makes no difference to the fixed-point multipliers as the only effect is on the position of the decimal point. As for the adders, this information means to perform data alignment by specifically extending the MSB (integer part) and LSB (fractional part) to certain numbers, which however does not incur any extra logic. On the other hand, the intra-layer consistency involves truncating the partial sum produced by the adder tree and tailoring the registered result to a target format. This inconsistency directly affects the truncator in size.

With the above model, the total size and latency of a single-layer accelerator can be approximated. As mentioned, the size of layer $l$ is a function of $Tn$, $Tm$, and bit width of weight and activations, incoming and outgoing, i.e.
\begin{equation}\label{eq:lut}
    Lut_i = qce(Tn_i, Tm_i, Ai_{i-1}, Af_{i-1}, Ai_i, Af_i, Wi_i, Wf_i),
\end{equation}
where $qce$ is the LUT approximator of the QCE for FPGAs and is predefined as a library. Besides, the latency of the single-layer accelerator can be  explicitly approximated by computation as
\begin{equation}\label{eq:lat}
    Lat_i \approx \left\lceil\frac{M_i}{Tm_i}\right\rceil\times
    \left\lceil\frac{N_i}{Tn_i}\right\rceil\times
    R_i\times C_i\times Fh_i\times Fw_i.
\end{equation}
Equation \eqref{eq:lut} and \eqref{eq:lat} are used to calculate the LUT usage and latency of a single-layer accelerator, upon which our multi-layer accelerator model is based. If a multi-layer partition $g$ contains consecutive layers from $i$ to $j$ operating in a pipelined fashion, then we have the overall size and latency as
\begin{equation}
    Lut_{g:i\sim j} = \sum\limits_{k=i}^{j}Lut_k,\  Lat_{g:i\sim j} = \max\limits_{i\le k\le j}Lat_k.
\end{equation}
Suppose a number of $G$ partitions covering a total of $L$ layers iterate their operations on the same FPGA, the total LUT usage and latency are then
\begin{equation}
    Lut_{1\sim G:1\sim L} = \max\limits_{1\le g\le G}Lut_{g:i\sim j},\ Lat_{1\sim G:1\sim L} = \sum\limits_{g=1}^{G}Lat_{g:i\sim j}
\end{equation}

\subsubsection{Searching Algorithm}
As implied by \eqref{eq:ispace}, the problem of searching the hardware space $\mathcal{H}$ involves deciding parameters $Tn\in[1, N]$, $Tm\in[1, N^\prime]$, as well as partitioning the $L$ layers into $G\in [1, L]$ clusters. Let $rL$ and $rT$ be the required LUT usage and throughput limits by the design specifications, we introduce $\mathbf{P}_L^{\langle rL, rT\rangle}$ to represent both this problem and solution set to this problem: given specification pair $\langle rL, rT\rangle$, $\mathbf{P}_L^{\langle rL, rT\rangle}$ returns all the possible solutions for implementing the CNN accelerator of $L$ layers. The task of our searching algorithm is to verify whether $\mathbf{P}_L^{\langle rL, rT\rangle}=\varnothing$.
In order to address this task, we also need to introduce the single-layer search problem as the basic tool: we use $\mathbf{p}_{l}^{\langle rL, rT\rangle}$ to represent the problem of searching for the hardware solutions to a single layer $l$ under the constraint of $\langle rL, rT\rangle$ and again its solution set. We shall solve the problem $\mathbf{P}_{L}^{\langle rL, rT\rangle}$ by incrementally solving the basic problem $\mathbf{p}_i^{\langle rL^{\prime}, rT^{\prime}\rangle}$, from $i=1$ to $L$.
\begin{algorithm}[h]
\caption{Dynamic Search in Hardware Space}
\label{alg:ds}
\begin{algorithmic}[1]
\REQUIRE $L$, $rL$, $rT$
\ENSURE solution set $\mathbf{S}$
\STATE Initialize $rl = rL$, $rt = rT$, \STATE Initialize $\mathbf{S}_{0}=\{s\}$ where $f_i(s)=0$ for $i$=1, 2, 3
\FOR{each $l = 1, 2, ..., L$}
    \FOR{each $s\in S_{l-1}$}
        \STATE compute $rl$ and $rt$ using Equation \eqref{eq:case1}
        \STATE compute $\mathbf{s}_{l} = F(\mathbf{p}_l^{\langle rl, rt\rangle})$
        \FORALL{$s^\prime \in \mathbf{s}_l$}
            \STATE append $s^\prime$ to the last partition of $s$ and add the result to $\mathbf{S}_{l}$
        \ENDFOR
        \STATE compute $rl$ and $rt$ using Equation \eqref{eq:case2}
        \STATE compute $\mathbf{s}_{l} = F(\mathbf{p}_l^{\langle rl, rt\rangle})$
        \FORALL{$s^{\prime\prime} \in \mathbf{s}_l$}
            \STATE set $s^{\prime\prime}$ as a new partition to $s$ and add the result to $\mathbf{S}_{l}$
        \ENDFOR
        \STATE update the frontier $\mathbf{S}_{l}=Fr(\mathbf{S}_{l})$
    \ENDFOR
\ENDFOR
\STATE $\mathbf{S}=\mathbf{S}_{L}$
\end{algorithmic}
\end{algorithm}

For any solution $s\in \mathbf{P}_L^{\langle rL, rT\rangle}$, we define three functions $f_1(s)$, $f_2(s)$ and $f_3(s)$ respectively as 1) the number of LUTs consumed by the last partition of $s$, 2) the overall latency of $s$, and 3) the sum of latency of all the partitions in $s$ except the last one. For any two solutions $s_1$ and $s_2$, if we have $f_1(s_1)\le f_1(s_2)$, $f_2(s_1)\le f_2(s_2)$, and $f_3(s_1)\le f_3(s_2)$, then $s_1$ is considered superior to $s_2$, and all the solutions that is not inferior to any other solution compose the Pareto Frontier of the solution set. Our algorithm is based on the fact that the existence of a solution is equivalent to the existence of the frontier of the solution set. Following this observation, we search for the frontier of $\mathbf{P}_L^{\langle rL, rT\rangle}$ such that the space is significantly pruned.  If the network has a depth of $l+1$, there are only two scenarios of how layer $(l+1)$ is related to the first $l$ layers:
\begin{enumerate}
    \item layer (l+1) is appended to the last partition of the previous $l$ layers, and
    \item layer (l+1) forms a partition itself.
\end{enumerate}
These two scenarios differ by the constraints to the problem of searching the last layer $\mathbf{p}_{l+1}^{\langle rl, rt\rangle}$. For every solution $s\in Fr(\mathbf{P}^{\langle rl, rt\rangle}_l)$, where $Fr(S)$ denotes the frontier of solution set $S$, we update the layer $l+1$ in both two manners under updated constraints. In the first case,
\begin{equation}\label{eq:case1}
    \left\{
    \begin{array}{ll}
        rl &= rL - f_1(s) \\
        rt &= (\frac{1}{rT} - \frac{f_3(s)}{clock\_rate})^{-1}
    \end{array} \right.
\end{equation}
and in the second case,
\begin{equation}\label{eq:case2}
    \left\{
    \begin{array}{ll}
        rl &= rL\\
        rt &= (\frac{1}{rT} - \frac{f_2(s)}{clock\_rate})^{-1}
    \end{array} \right.
\end{equation}
When $\mathbf{p}_{l+1}^{\langle rl, rt\rangle}$ is solved, it would also be  sufficient to keep only its frontier with respect to only $f_1$ and $f_2$. Then $Fr(\mathbf{p}_{l+1}^{rl, rt})$ is achieved by combining $Fr(\mathbf{P}^{\langle rl^{\prime}, rt^{\prime}\rangle}_l)$ and $F(\mathbf{p}_{l+1}^{\langle rl, rt\rangle})$ in the corresponding way of how layer $l+1$ is derived. The full procedure is shown in Algorithm \ref{alg:ds}.

\section{Experiment}\label{sec:exp}
We apply the joint architeture and quantization search method to design CNN accelerators. The design objective is the classification task on CIFAR-10 dataset that satisfies two constraints of available LUTs and throughput tolerance. This dataset provides 50000 images for training and 10000 for testing, and the entirety of them are used in the search process. For the training set, augmentation techniques are applied for tuning a network which consists of normalization, rotation, shifting, and random flip.

\begin{table}[!htb]
  \caption{The architecture and quantization space of each CNN layer used in the experiment}
  \label{tab:space}
  \begin{tabular}{ccc}
    \toprule
    Parameter & Symbol & Value\\
    \midrule
    \# filters & $N$ & (24, 36, 48, 64)\\
    filter height & $Fh$ & (1, 3, 5, 7)\\
    filter width & $Fw$ & (1, 3, 5, 7)\\
    stride height & $Sh$ & (1, 2, 3)\\
    stride width & $Sw$ & (1, 2, 3)\\
    pooling size & $Ps$ & (1, 2)\\
    activation integer bits & $Ai$ & (0, 1, 2, 3)\\
    activation fractional bits & $Af$ & (0, 1, 2, 3, 4, 5, 6)\\
    weight integer bits & $Wi$ & (0, 1, 2, 3)\\
    weight fractional bits & $Wf$ & (0, 1, 2, 3, 4, 5, 6)\\
  \bottomrule
\end{tabular}
\end{table}

\begin{table*}[!htb]
  \caption{Architectural information of the sampled designs. A\textsubscript{1} and A\textsubscript{2} are the best architectures found by NAS in 1000 episodes and their layered hyper-parameters are given in the form of (N, Fh, Fw, Sh, Sw, Ps). Then we remove the strides from the search and get B\textsubscript{1} and B\textsubscript{2} whose parameters of each layer are listed as (N, Fh, Fw, Ps). D, E, and F are the results from our joint search process with the same architecture space but no strides.}
  \label{tab:architecture}
  \begin{tabular}{cccccccccc}
    \toprule
     Network & Layer 1 & Layer 2 & Layer 3 & Layer 4 & Layer 5 & Layer 6 & \#paras & Acc w/o BN & Acc w/ BN\\
    \midrule
    A\textsubscript{1}& (64,3,3,1,1,1)&(48,7,5,1,1,1)&(48,5,5,2,1,1)&(64,3,5,1,1,1)&(36,5,7,1,1,1)&(64,3,1,1,2,2)&
    300,804 & 87.76\% & 88.96\% \\
    A\textsubscript{2}&
    (24,3,3,1,1,1)&(36,5,5,1,1,1)&(64,5,5,2,1,1)&(64,5,5,1,1,1)&(24,5,5,1,2,1)&(64,3,3,1,2,1)&
    234,748 & 87.46\% & 88.87\% \\
    B\textsubscript{1}&
    (64,3,3,1)&(64,3,5,1)&(64,3,3,2)&(64,5,5,2)&(64,5,3,1)&(64,7,7,1)&
    464,960 & 89.71\% & 90.30\% \\
    B\textsubscript{2} &
    (64,5,3,1)&(64,3,5,1)&(64,3,5,2)&(64,5,5,2)&(64,5,3,1)&(64,7,7,1)&
    490,688 & 89.38\% & 90.49\% \\
    \hline
    D & (48,5,3,1) & (48,3,1,2) & (36,1,7,2) & (36,7,3,1) & (24,5,5,1) & (24,1,1,1) & 70,776 & 83.65\% & 84.31\% \\
    E & (48,5,1,1) & (48,5,3,2) & (36,1,5,1) & (64,7,7,2) & (64,7,3,2) & (48,5,3,1) & 289,220 & 86.99\% & 88.27\% \\
    F & (64,1,5,1) & (36,1,7,1) & (64,5,7,2) & (48,5,3,2) & (48,7,7,1) & (36,1,5,1) & 26,5640 & 87.03\% & 88.42\% \\
    \bottomrule
\end{tabular}
\end{table*}
% \begin{table*}[!htb]
%   \caption{Implementation information of the sampled designs. For network A and B, the designs are found by quantization search to certain architectures in Table~\ref{tab:architecture}. For D, E and F, the quantization and implementation on hardware are designed together with their architectures. The quantization details are shown in Figure \ref{tab:quantization}.}
%   \label{tab:design}
%   \begin{tabular}{cccccccc}
%     \toprule
%      Design & rL & rT & Acc & Acc & \#LUTs & FPS & para size\\
%      & & & w/o Quan & w/ Quan & & & (kbits) \\
%     \midrule
%     A\textsubscript{1}-d\textsubscript{1} & 100,000 & 500 & \multirow{1}{*}{87.76\%} & 80.23\% & 99,871 & 556 & 1,867 \\
%      A\textsubscript{1}-d\textsubscript{2} & 100,000 & 1000 & 87.76\% & 25.79\% & 99,848 & 1157 & 1,189 \\
%     B\textsubscript{1}-d\textsubscript{1} & 100,000 & 500 & \multirow{1}{*}{89.71\%} & 87.64\% & 96,904 & 512 & 3,463 \\
%     B\textsubscript{1}-d\textsubscript{2} & 100,000 & 1000 & 89.71\% & 64.35\% & 98,752 & 1020 & 2,784 \\
%     B\textsubscript{1}-d\textsubscript{3} & 300,000 & 2000 & 89.71\% & 50.93\% & 285,441 & 2083 & 2,835 \\
%     \hline
%     D & \multirow{1}{*}{30,000}&1000&83.65\%&82.98\%&29,904&1293&457\\
%     E & \multirow{1}{*}{100,000}&1000&86.99\%&82.76\%&94,496&1042&1,923\\
%     F & \multirow{1}{*}{300,000}&2000&87.03\%&84.92\%&299,860&2089&1,217\\
%     \bottomrule
% \end{tabular}
% \end{table*}

\begin{table*}[!htb]
  \caption{Implementation information of the sampled designs. For network A and B, the designs are found by quantization search to certain architectures in Table~\ref{tab:architecture}. For D, E and F, the quantization and implementation on hardware are designed together with their architectures. The quantization details are shown in Figure \ref{tab:quantization}.}
  \label{tab:design}
  \begin{tabular}{cccccccc}
    \toprule
     Design & rL & rT & Acc w/o quantization & Acc w/ quantization & \#LUTs & Throughput (frames/s)& parameter size (kbits)\\
    \midrule
    A\textsubscript{1}-d\textsubscript{1} & 100,000 & 500 & \multirow{1}{*}{87.76\%} & 80.23\% & 99,871 & 556 & 1,867 \\
     A\textsubscript{1}-d\textsubscript{2} & 100,000 & 1000 & 87.76\% & 25.79\% & 99,848 & 1157 & 1,189 \\
    B\textsubscript{1}-d\textsubscript{1} & 100,000 & 500 & \multirow{1}{*}{89.71\%} & 87.64\% & 96,904 & 512 & 3,463 \\
    B\textsubscript{1}-d\textsubscript{2} & 100,000 & 1000 & 89.71\% & 64.35\% & 98,752 & 1020 & 2,784 \\
    B\textsubscript{1}-d\textsubscript{3} & 300,000 & 2000 & 89.71\% & 50.93\% & 285,441 & 2083 & 2,835 \\
    \hline
    D & \multirow{1}{*}{30,000}&1000&83.65\%&82.98\%&29,904&1293&457\\
    E & \multirow{1}{*}{100,000}&1000&86.99\%&82.76\%&94,496&1042&1,923\\
    F & \multirow{1}{*}{300,000}&2000&87.03\%&84.92\%&299,860&2089&1,217\\
    \bottomrule
\end{tabular}
\end{table*}

\begin{figure}
     \centering
     \begin{subfigure}[b]{0.22\textwidth}
         \centering
         \includegraphics[width=\textwidth]{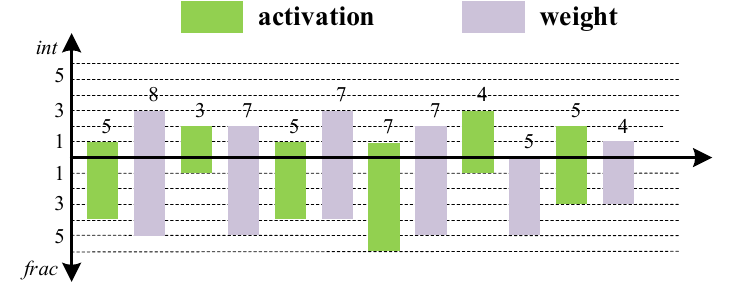}
         \caption{A\textsubscript{1}-d\textsubscript{1}}
         \label{fig:A1d1}
     \end{subfigure}
     \hfill
     \begin{subfigure}[b]{0.22\textwidth}
         \centering
         \includegraphics[width=\textwidth]{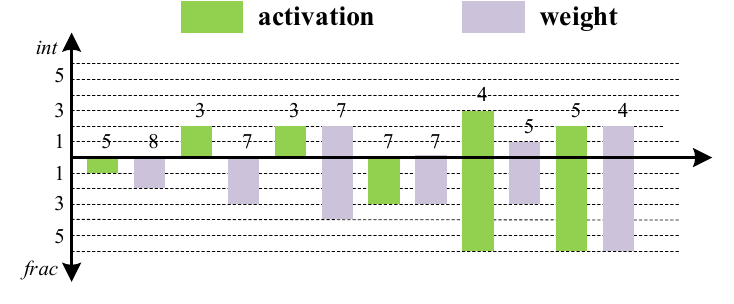}
         \caption{A\textsubscript{1}-d\textsubscript{2}}
         \label{fig:A1d2}
     \end{subfigure}
     \begin{subfigure}[b]{0.22\textwidth}
         \centering
         \includegraphics[width=\textwidth]{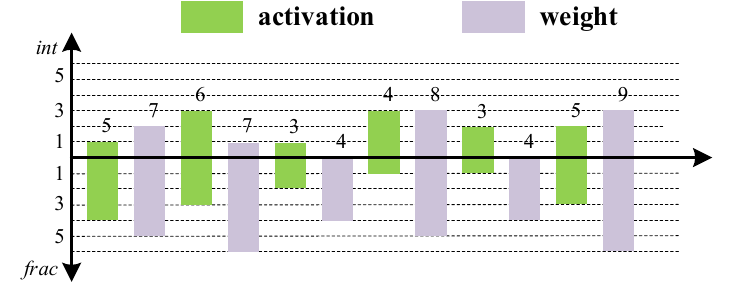}
         \caption{B\textsubscript{1}-d\textsubscript{1}}
         \label{fig:B1d1}
     \end{subfigure}
     \hfill
     \begin{subfigure}[b]{0.22\textwidth}
         \centering
         \includegraphics[width=\textwidth]{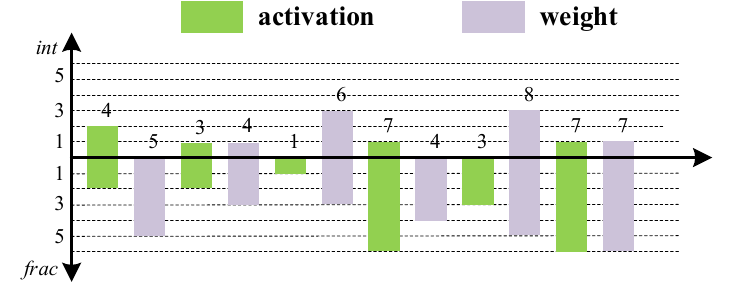}
         \caption{B\textsubscript{1}-d\textsubscript{2}}
         \label{fig:B1d2}
     \end{subfigure}
     \begin{subfigure}[b]{0.22\textwidth}
         \centering
         \includegraphics[width=\textwidth]{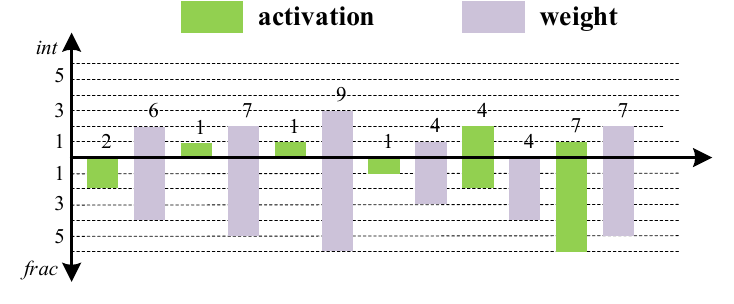}
         \caption{B\textsubscript{1}-d\textsubscript{3}}
         \label{fig:B1d3}
     \end{subfigure}
     \hfill
     \begin{subfigure}[b]{0.22\textwidth}
         \centering
         \includegraphics[width=\textwidth]{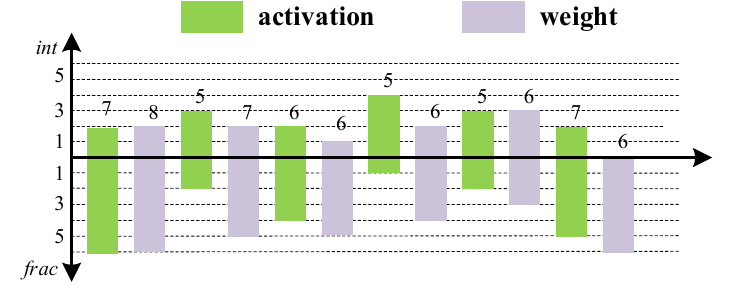}
         \caption{D}
         \label{fig:D}
     \end{subfigure}
     \begin{subfigure}[b]{0.22\textwidth}
         \centering
         \includegraphics[width=\textwidth]{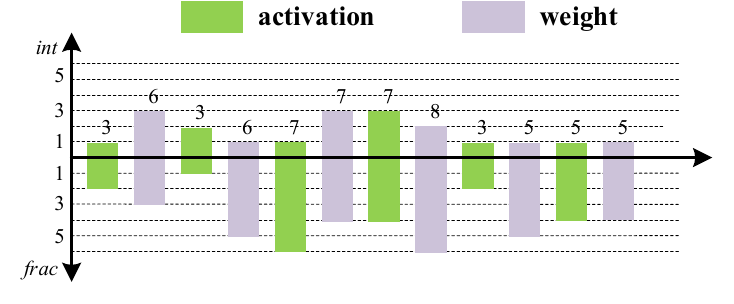}
         \caption{E}
         \label{fig:E}
     \end{subfigure}
     \hfill
     \begin{subfigure}[b]{0.22\textwidth}
         \centering
         \includegraphics[width=\textwidth]{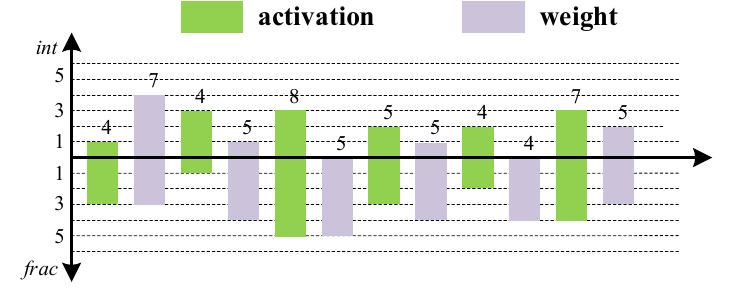}
         \caption{F}
         \label{fig:F}
     \end{subfigure}
     \vspace{-10pt}
        \caption{Quantization details of the sampled designs}
        \label{fig:quantization}
\end{figure}
The architecture and quantization space used in the experiment is listed in Table~\ref{tab:space}. For the child network, we assume each layer is composed of certainly a convolutional operation followed by rectified linear units, and possibly zero-padding and max-pooling operations before and after it. After the convolutional layers, two fully connected layers are tailing to output the prediction distributions, which are not included as part in our hardware model. We train the child CNN for 30 epochs with Stochastic Gradient Decent (SGD) algorithm taking batches of 128 images, learning rate of 0.01, and momentum of 0.9. Once the training is finished, the reward is the test accuracy averaged over the last 5 epochs. After the search, the best sample are selected and tuned for 150 epochs, along with 64 batch size and decaying learning rates from 0.01 downward to 0.0001. The highest accuracy along the tuning process is what is finally reported. On the other side, the controller consists of a two-layer LSTM cell with 35 hidden units at each layer accompanied with an embedding and fully-connected layer at each time step of corresponding dimensions. To train the controller, we apply the Stochastic Gradient Ascent (SGA) algorithm with a learning rate of 0.2 and batch size of 5. The baseline is the exponential moving average of the previous rewards.
Finally, we build the hardware model based upon an Altera Cyclone IV FPGA platform where we set the global clock rate as 100 MHz. In order to be build practical FPGA synthesis, we set the depth of the child network to have 6 layers, and designate the allowable LUT usage at three scales: 30,000, 100,000, and 300,000.

For comparison, we first perform the NAS to find architectures with good performances in accuracy and then search the best quantization to fit them under some hardware specifications. The sampled networks with highest accuracy on the test set and their best design with quantization are shown in Table \ref{tab:architecture} and \ref{tab:quantization}, respectively. We use the floating-point accuracy in Table \ref{tab:architecture} as the reference of our design of quantization. Note that the same architecture is about 1\% less performant without the batch normalization (BN), but that is just what we shall refer to. The reason is twofold: 1) the BN involves operations that require additional hardware support for both computation and memory, and more importantly 2) it will make the quantized network extremely unstable whose output may have a prohibitively large variance for the searching algorithm. Next, we use the joint method to search architecture and quantization together with the specifications under which the quantization search fails to provide a good result in a general sense with respect to the best architectures found. As a result, we have three designs of the CNN accelerators with different hardware specifications. The details of these designs are reported in Table \ref{tab:design}.

\begin{table}[!htb]
  \caption{Quantization search result for the sampled networks A and B in Table \ref{tab:architecture}. The best accuracy on the test set of CIFAR10 in 2000 episodes are reported.}
  \vspace{-10pt}
  \label{tab:quantization}
  \begin{tabular}{ccccc}
    \toprule
     Network & rT & rL=30,000 & rL=100,000 & rL=300,000\\
    \midrule
    \multirow{3}{*}{A\textsubscript{1}}&500&10.65\%&80.23\%&86.16\%\\
    &1000&x&25.79\%&84.90\%\\
    &2000&x&x&x\\
    \hline
    \multirow{3}{*}{A\textsubscript{2}}&500& 55.45\% & 85.92\% & 86.26\% \\
    &1000& x & 67.30\% & 76.51\% \\
    &2000 & x & x & x\\
    \hline
    \multirow{3}{*}{B\textsubscript{1}}&500&10.02\%&87.64\%&87.34\%\\
    & 1000 & x & 64.35\% & 87.43\% \\
    &2000& x & x & 50.93\% \\
    \hline
    \multirow{3}{*}{B\textsubscript{2}}&500 & 10.20\% & 85.31\% & 88.53\% \\
    &1000 & x & 43.81\% & 86.50\% \\
    & 2000 & x & x & 16.71\% \\
    \bottomrule
\end{tabular}
\end{table}

\begin{figure}
     \centering
     \begin{subfigure}[b]{0.22\textwidth}
         \centering
         \includegraphics[width=\textwidth]{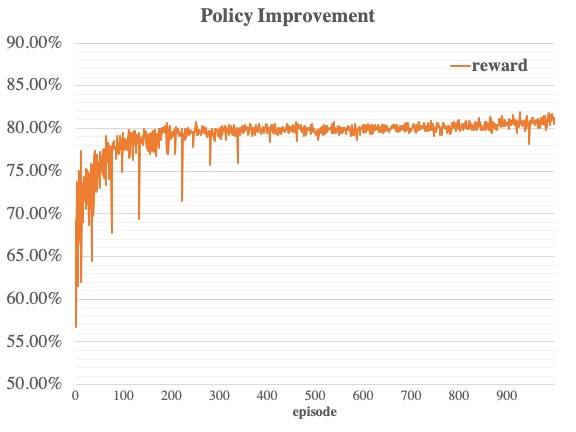}
         \caption{NAS with stride}
         \label{fig:nas_training}
     \end{subfigure}
     \hfill
     \begin{subfigure}[b]{0.22\textwidth}
         \centering
         \includegraphics[width=\textwidth]{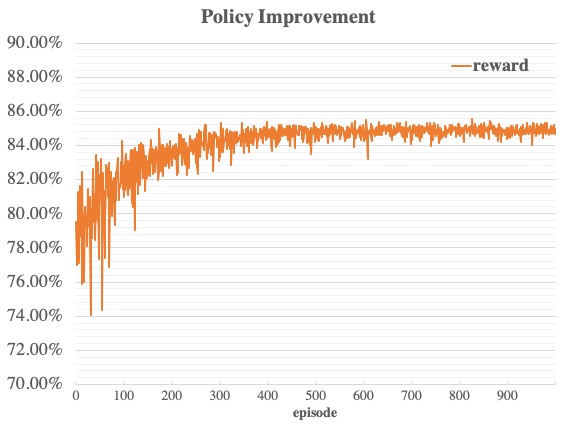}
         \caption{NAS without stride}
         \label{fig:nas_training_ns}
     \end{subfigure}
     \begin{subfigure}[b]{0.22\textwidth}
         \centering
         \includegraphics[width=\textwidth]{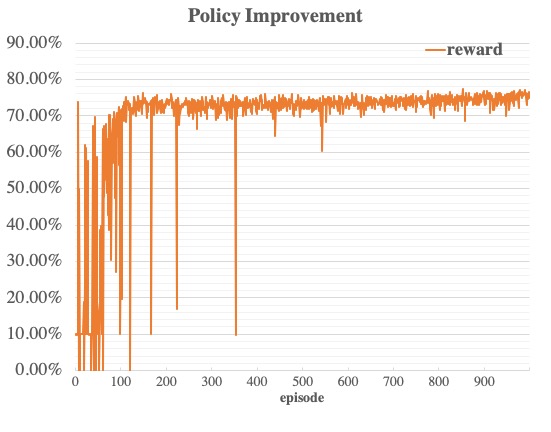}
         \caption{Joint: rL=30000, rT=500}
         \label{fig:joint_30000}
     \end{subfigure}
     \hfill
     \begin{subfigure}[b]{0.22\textwidth}
         \centering
         \includegraphics[width=\textwidth]{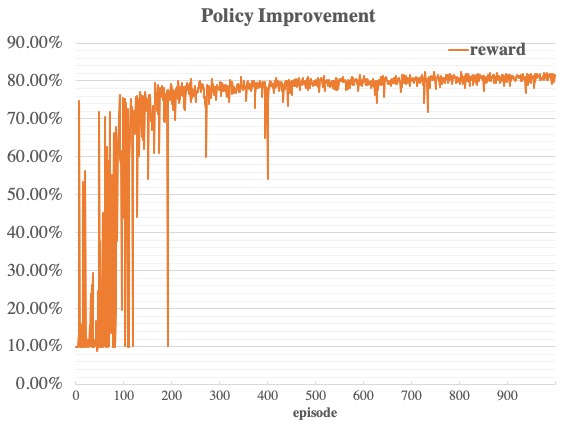}
         \caption{Joint: rL=300000, rT=500}
         \label{fig:joint_300000}
     \end{subfigure}
     \vspace{-10pt}
        \caption{The plotted training process: comparison between  NAS and Joint search.}
        \label{fig:training}
\end{figure}

Table \ref{tab:quantization} shows the quantization search results with throughput requirement of 500, 1000, and 2000 frames per second. It is clearly noted the drop in accuracy with increasing throughput is very sharp. For example, the accuracy of the quantized network B\textsubscript{1} drops from 87.64\% to 64.35\% with doubling 500 MHz throughput requirement at rL=100,100. It is implied although the architecture has an excellent original performance, it is too sensitive to quantization to be suitable for resource-constrained hardware design. On the other hand, our joint search method has found a solution to achieve 82.76\% accuracy and meanwhile the 1000 throughput. In contrast, the original accuracy of architecture E is only 86.99\%, worse than B\textsubscript{1} by nearly 3\%, but the its accuracy is more robust to quantization with 4\% degradation. With the joint search, we could even found a design using less than 30,000 LUTs but achieving 82.98\% accuracy and 1293 throughput. Note there are no valid designs for almost every sample architecture even with 300,000 available LUTs.

We further compare the quantization designs for A\textsubscript{1} and B\textsubscript{1} with those using joint search for D, E, and F. As illustrated in Figure \ref{fig:quantization}, the convolutional layers generally exhibit different patterns in terms of bit-width requirement. Another observation is with fixed architecture, the quantization search tends to spend more bits on the weight but not the activations, while the joint search treats these two values fairly.

\section{Conclusion and Challenge}\label{sec:con}
In this paper, we overviewed the recent development of automatic machine learning, identifying the trend towards the hardware-software co-design using NAS. A hardware-aware co-design framework is proposed to jointly explore architecture, quantization, and hardware design space. It is proved by experiment the joint search can provides much more flexibility in compressed design robust performance as compared to the traditional artificial design using fixed architecture.

In this project, however, the existence of difficulties in applying hardware-aware NAS is also identified. Compared to pure NAS, the controller is forced the burden to learn the hardware constraint from the beginning of the search process, resulting in a higher variance and early convergence to local optimality (Figure \ref{fig:training}). On the other hand, the search process is computation-intensive and resource-/time-consuming. Lastly, the hardware exploration and design automation heavily rely on the hardware model that needs to be built with more effort. There remains a lot of room of improvement in this topic.

\section*{Acknowledgement}
This work was supported in part by the National
Science Foundation under Grant CCF-1820537 and CNS-1822099.

\vspace{-10pt}


%\bibliographystyle{ACM-Reference-Format}
%\bibliography{sample-base}

\begin{thebibliography}{27}

%%% the Web address.

\ifx \showCODEN    \undefined \def \showCODEN     #1{\unskip}     \fi
\ifx \showDOI      \undefined \def \showDOI       #1{#1}\fi
\ifx \showISBNx    \undefined \def \showISBNx     #1{\unskip}     \fi
\ifx \showISBNxiii \undefined \def \showISBNxiii  #1{\unskip}     \fi
\ifx \showISSN     \undefined \def \showISSN      #1{\unskip}     \fi
\ifx \showLCCN     \undefined \def \showLCCN      #1{\unskip}     \fi
\ifx \shownote     \undefined \def \shownote      #1{#1}          \fi
\ifx \showarticletitle \undefined \def \showarticletitle #1{#1}   \fi
\ifx \showURL      \undefined \def \showURL       {\relax}        \fi
% invisible to TeX
\providecommand\bibfield[2]{#2}
\providecommand\bibinfo[2]{#2}
\providecommand\natexlab[1]{#1}
\providecommand\showeprint[2][]{arXiv:#2}

\bibitem[\protect\citeauthoryear{Bender et~al\mbox{.}}{Bender
  et~al\mbox{.}}{2018}]%
        {bender2018understanding}
\bibfield{author}{\bibinfo{person}{Gabriel Bender} {et~al\mbox{.}}}
  \bibinfo{year}{2018}\natexlab{}.
\newblock \showarticletitle{Understanding and simplifying one-shot architecture
  search}. In \bibinfo{booktitle}{\emph{Int. Conf. on Machine Learning}}.
  \bibinfo{pages}{549--558}.
\newblock


\bibitem[\protect\citeauthoryear{Cai et~al\mbox{.}}{Cai et~al\mbox{.}}{2018}]%
        {cai2018proxylessnas}
\bibfield{author}{\bibinfo{person}{Han Cai} {et~al\mbox{.}}}
  \bibinfo{year}{2018}\natexlab{}.
\newblock \showarticletitle{ProxylessNAS: Direct neural architecture search on
  target task and hardware}.
\newblock \bibinfo{journal}{\emph{arXiv preprint arXiv:1812.00332}}
  (\bibinfo{year}{2018}).
\newblock


\bibitem[\protect\citeauthoryear{Jiang et~al\mbox{.}}{Jiang
  et~al\mbox{.}}{2016}]%
        {jiang2016optimal}
\bibfield{author}{\bibinfo{person}{Weiwen Jiang} {et~al\mbox{.}}}
  \bibinfo{year}{2016}\natexlab{}.
\newblock \showarticletitle{Optimal functional-unit assignment and buffer
  placement for probabilistic pipelines}. In \bibinfo{booktitle}{\emph{2016
  Int. Conf. on Hardware/Software Codesign and System Synthesis (CODES+
  ISSS)}}. IEEE, \bibinfo{pages}{1--10}.
\newblock


\bibitem[\protect\citeauthoryear{Jiang et~al\mbox{.}}{Jiang
  et~al\mbox{.}}{2017}]%
        {jiang2017optimal}
\bibfield{author}{\bibinfo{person}{Weiwen Jiang} {et~al\mbox{.}}}
  \bibinfo{year}{2017}\natexlab{}.
\newblock \showarticletitle{Optimal functional unit assignment and voltage
  selection for pipelined MPSoC with guaranteed probability on time
  performance}. In \bibinfo{booktitle}{\emph{ACM SIGPLAN Notices}},
  Vol.~\bibinfo{volume}{52}. ACM, \bibinfo{pages}{41--50}.
\newblock


\bibitem[\protect\citeauthoryear{Jiang et~al\mbox{.}}{Jiang
  et~al\mbox{.}}{2018}]%
        {jiang2018heterogeneous}
\bibfield{author}{\bibinfo{person}{Weiwen Jiang} {et~al\mbox{.}}}
  \bibinfo{year}{2018}\natexlab{}.
\newblock \showarticletitle{Heterogeneous fpga-based cost-optimal design for
  timing-constrained cnns}.
\newblock \bibinfo{journal}{\emph{IEEE Trans. Comput.-Aided Design of Integr.
  Circuits and Syst}} \bibinfo{volume}{37}, \bibinfo{number}{11}
  (\bibinfo{year}{2018}), \bibinfo{pages}{2542--2554}.
\newblock


\bibitem[\protect\citeauthoryear{Jiang et~al\mbox{.}}{Jiang
  et~al\mbox{.}}{2019a}]%
        {jiang2019accuracy}
\bibfield{author}{\bibinfo{person}{Weiwen Jiang} {et~al\mbox{.}}}
  \bibinfo{year}{2019}\natexlab{a}.
\newblock \showarticletitle{Accuracy vs. Efficiency: Achieving Both through
  FPGA-Implementation Aware Neural Architecture Search}. In
  \bibinfo{booktitle}{\emph{Proc. 56th Annual Design Automation Conference
  2019}}. ACM, \bibinfo{pages}{5}.
\newblock


\bibitem[\protect\citeauthoryear{Jiang et~al\mbox{.}}{Jiang
  et~al\mbox{.}}{2019b}]%
        {jiang2019hardware}
\bibfield{author}{\bibinfo{person}{Weiwen Jiang} {et~al\mbox{.}}}
  \bibinfo{year}{2019}\natexlab{b}.
\newblock \showarticletitle{Hardware/software co-exploration of neural
  architectures}.
\newblock \bibinfo{journal}{\emph{arXiv preprint arXiv:1907.04650}}
  (\bibinfo{year}{2019}).
\newblock


\bibitem[\protect\citeauthoryear{Jiang et~al\mbox{.}}{Jiang
  et~al\mbox{.}}{2019c}]%
        {jiang2019xfer}
\bibfield{author}{\bibinfo{person}{Weiwen Jiang} {et~al\mbox{.}}}
  \bibinfo{year}{2019}\natexlab{c}.
\newblock \showarticletitle{XFER: a novel design to achieve super-linear
  performance on multiple FPGAs for real-time AI}. In
  \bibinfo{booktitle}{\emph{Proc. of Int. Symp. on FPGA}}. ACM,
  \bibinfo{pages}{305--305}.
\newblock


\bibitem[\protect\citeauthoryear{Koeplinger et~al\mbox{.}}{Koeplinger
  et~al\mbox{.}}{2016}]%
        {koeplinger2016automatic}
\bibfield{author}{\bibinfo{person}{David Koeplinger} {et~al\mbox{.}}}
  \bibinfo{year}{2016}\natexlab{}.
\newblock \showarticletitle{Automatic generation of efficient accelerators for
  reconfigurable hardware}. In \bibinfo{booktitle}{\emph{2016 ACM/IEEE 43rd
  Annual Int. Symp. on Comput. Archit. (ISCA)}}. IEEE,
  \bibinfo{pages}{115--127}.
\newblock


\bibitem[\protect\citeauthoryear{Li et~al\mbox{.}}{Li et~al\mbox{.}}{2019}]%
        {li2019exploiting}
\bibfield{author}{\bibinfo{person}{Boyang Li} {et~al\mbox{.}}}
  \bibinfo{year}{2019}\natexlab{}.
\newblock \showarticletitle{Exploiting computation power of blockchain for
  biomedical image segmentation}. In \bibinfo{booktitle}{\emph{IEEE Conf. on
  Computer Vision and Pattern Recognition Workshops}}.
\newblock


\bibitem[\protect\citeauthoryear{Liu, Simonyan, and Yang}{Liu
  et~al\mbox{.}}{2018}]%
        {liu2018darts}
\bibfield{author}{\bibinfo{person}{Hanxiao Liu}, \bibinfo{person}{Karen
  Simonyan}, {and} \bibinfo{person}{Yiming Yang}.}
  \bibinfo{year}{2018}\natexlab{}.
\newblock \showarticletitle{Darts: differentiable architecture search}.
\newblock \bibinfo{journal}{\emph{arXiv preprint arXiv:1806.09055}}
  (\bibinfo{year}{2018}).
\newblock


\bibitem[\protect\citeauthoryear{Research and Markets}{Research and
  Markets}{2018}]%
        {aiot}
\bibfield{author}{\bibinfo{person}{Research} {and} \bibinfo{person}{Markets}.}
  \bibinfo{year}{2018}\natexlab{}.
\newblock \showarticletitle{Artificial intelligence in IoT: AIoT Technology,
  Platforms, Applications and Services by Industry Vertical 2018 - 2023}.
\newblock \bibinfo{journal}{\emph{Report}} (\bibinfo{year}{2018}).
\newblock


\bibitem[\protect\citeauthoryear{Tan et~al\mbox{.}}{Tan et~al\mbox{.}}{2018}]%
        {tan2018mnasnet}
\bibfield{author}{\bibinfo{person}{Mingxing Tan} {et~al\mbox{.}}}
  \bibinfo{year}{2018}\natexlab{}.
\newblock \showarticletitle{Mnasnet: Platform-aware neural architecture search
  for mobile}.
\newblock \bibinfo{journal}{\emph{arXiv preprint arXiv:1807.11626}}
  (\bibinfo{year}{2018}).
\newblock


\bibitem[\protect\citeauthoryear{Wang et~al\mbox{.}}{Wang
  et~al\mbox{.}}{2019a}]%
        {wang2019msu}
\bibfield{author}{\bibinfo{person}{Tianchen Wang} {et~al\mbox{.}}}
  \bibinfo{year}{2019}\natexlab{a}.
\newblock \showarticletitle{MSU-Net: multiscale statistical U-Net for real-time
  3D cardiac MRI video segmentation}. In \bibinfo{booktitle}{\emph{Proc. of
  Medical Image Computing and Computer Assisted Interventions (MICCAI)}}.
  \bibinfo{pages}{0--0}.
\newblock


\bibitem[\protect\citeauthoryear{Wang et~al\mbox{.}}{Wang
  et~al\mbox{.}}{2019b}]%
        {tianchen2019}
\bibfield{author}{\bibinfo{person}{Tianchen Wang} {et~al\mbox{.}}}
  \bibinfo{year}{2019}\natexlab{b}.
\newblock \showarticletitle{SCNN: a general distribution based statistical
  convolutional neural network with application to video object detection}. In
  \bibinfo{booktitle}{\emph{AAAI Conf. on AI}}.
\newblock


\bibitem[\protect\citeauthoryear{Williams}{Williams}{1992}]%
        {WILLIAMS1992Simple}
\bibfield{author}{\bibinfo{person}{Ronald~J Williams}.}
  \bibinfo{year}{1992}\natexlab{}.
\newblock \showarticletitle{Simple statistical gradient-following algorithms
  for connectionist reinforcement learning}.
\newblock \bibinfo{journal}{\emph{Machine learning}} \bibinfo{volume}{8},
  \bibinfo{number}{3-4} (\bibinfo{year}{1992}), \bibinfo{pages}{229--256}.
\newblock


\bibitem[\protect\citeauthoryear{Wu et~al\mbox{.}}{Wu et~al\mbox{.}}{2018}]%
        {wu2018FBNet}
\bibfield{author}{\bibinfo{person}{Bichen Wu} {et~al\mbox{.}}}
  \bibinfo{year}{2018}\natexlab{}.
\newblock \showarticletitle{FBNet: hardware-qware efficient convNet design via
  differentiable neural architecture search}.
\newblock \bibinfo{journal}{\emph{arXiv preprint arXiv:1812.03443}}
  (\bibinfo{year}{2018}).
\newblock


\bibitem[\protect\citeauthoryear{Xu et~al\mbox{.}}{Xu et~al\mbox{.}}{2017}]%
        {xu2017edge}
\bibfield{author}{\bibinfo{person}{Xiaowei Xu} {et~al\mbox{.}}}
  \bibinfo{year}{2017}\natexlab{}.
\newblock \showarticletitle{Edge segmentation: empowering mobile telemedicine
  with compressed cellular neural networks}. In \bibinfo{booktitle}{\emph{Proc.
  of the 36th Int. Conf. on Computer-Aided Design}}. IEEE Press,
  \bibinfo{pages}{880--887}.
\newblock


\bibitem[\protect\citeauthoryear{Xu et~al\mbox{.}}{Xu et~al\mbox{.}}{2018a}]%
        {xu2018efficient}
\bibfield{author}{\bibinfo{person}{Xiaowei Xu} {et~al\mbox{.}}}
  \bibinfo{year}{2018}\natexlab{a}.
\newblock \showarticletitle{Efficient hardware implementation of cellular
  neural networks with incremental quantization and early exit}.
\newblock \bibinfo{journal}{\emph{ACM J. on Emerging Technologies in Computing
  Systems (JETC)}} \bibinfo{volume}{14}, \bibinfo{number}{4}
  (\bibinfo{year}{2018}), \bibinfo{pages}{48}.
\newblock


\bibitem[\protect\citeauthoryear{Xu et~al\mbox{.}}{Xu et~al\mbox{.}}{2018b}]%
        {xu2018quantization}
\bibfield{author}{\bibinfo{person}{Xiaowei Xu} {et~al\mbox{.}}}
  \bibinfo{year}{2018}\natexlab{b}.
\newblock \showarticletitle{Quantization of fully convolutional networks for
  accurate biomedical image segmentation}.
\newblock \bibinfo{journal}{\emph{Preprint at https://arxiv.
  org/abs/1803.04907}} (\bibinfo{year}{2018}).
\newblock


\bibitem[\protect\citeauthoryear{Xu et~al\mbox{.}}{Xu et~al\mbox{.}}{2018c}]%
        {xu2018resource}
\bibfield{author}{\bibinfo{person}{Xiaowei Xu} {et~al\mbox{.}}}
  \bibinfo{year}{2018}\natexlab{c}.
\newblock \showarticletitle{Resource constrained cellular neural networks for
  real-time obstacle detection using FPGAs}. In \bibinfo{booktitle}{\emph{2018
  19th Int. Symp. on Quality Electronic Design}}. IEEE,
  \bibinfo{pages}{437--440}.
\newblock


\bibitem[\protect\citeauthoryear{Xu et~al\mbox{.}}{Xu et~al\mbox{.}}{2018d}]%
        {xu2018scaling}
\bibfield{author}{\bibinfo{person}{Xiaowei Xu} {et~al\mbox{.}}}
  \bibinfo{year}{2018}\natexlab{d}.
\newblock \showarticletitle{Scaling for edge inference of deep neural
  networks}.
\newblock \bibinfo{journal}{\emph{Nature Electronics}} \bibinfo{volume}{1},
  \bibinfo{number}{4} (\bibinfo{year}{2018}), \bibinfo{pages}{216}.
\newblock


\bibitem[\protect\citeauthoryear{Xu et~al\mbox{.}}{Xu et~al\mbox{.}}{2019}]%
        {xu2019whole}
\bibfield{author}{\bibinfo{person}{Xiaowei Xu} {et~al\mbox{.}}}
  \bibinfo{year}{2019}\natexlab{}.
\newblock \showarticletitle{Whole-heart and great vessel segmentation in
  congenital heart disease using deep neural networks and graph matching}. In
  \bibinfo{booktitle}{\emph{Proc. of Medical Image Computing and Computer
  Assisted Interventions (MICCAI)}}. \bibinfo{pages}{0--0}.
\newblock


\bibitem[\protect\citeauthoryear{Zhang et~al\mbox{.}}{Zhang
  et~al\mbox{.}}{2015}]%
        {Zhang:2015:OFA:2684746.2689060}
\bibfield{author}{\bibinfo{person}{Chen Zhang} {et~al\mbox{.}}}
  \bibinfo{year}{2015}\natexlab{}.
\newblock \showarticletitle{Optimizing fpga-based accelerator design for deep
  convolutional neural networks}. In \bibinfo{booktitle}{\emph{Proc. of FPGA}}.
  ACM, \bibinfo{pages}{161--170}.
\newblock


\bibitem[\protect\citeauthoryear{Zhang et~al\mbox{.}}{Zhang
  et~al\mbox{.}}{2019}]%
        {ZhangJSH19}
\bibfield{author}{\bibinfo{person}{Xinyi Zhang} {et~al\mbox{.}}}
  \bibinfo{year}{2019}\natexlab{}.
\newblock \showarticletitle{When Neural Architecture Search Meets Hardware
  Implementation: from Hardware Awareness to Co-Design}. In
  \bibinfo{booktitle}{\emph{Proc. of ISLVLSI}}. \bibinfo{pages}{25--30}.
\newblock


\bibitem[\protect\citeauthoryear{Zoph et~al\mbox{.}}{Zoph
  et~al\mbox{.}}{2016}]%
        {zoph2016neural}
\bibfield{author}{\bibinfo{person}{Barret Zoph} {et~al\mbox{.}}}
  \bibinfo{year}{2016}\natexlab{}.
\newblock \showarticletitle{Neural architecture search with reinforcement
  learning}.
\newblock \bibinfo{journal}{\emph{arXiv preprint arXiv:1611.01578}}
  (\bibinfo{year}{2016}).
\newblock


\bibitem[\protect\citeauthoryear{Zoph et~al\mbox{.}}{Zoph
  et~al\mbox{.}}{2018}]%
        {zoph2018learning}
\bibfield{author}{\bibinfo{person}{Barret Zoph} {et~al\mbox{.}}}
  \bibinfo{year}{2018}\natexlab{}.
\newblock \showarticletitle{Learning transferable architectures for scalable
  image recognition}. In \bibinfo{booktitle}{\emph{Proceedings of the IEEE
  Conf. on computer vision and pattern recognition}}.
  \bibinfo{pages}{8697--8710}.
\newblock


\end{thebibliography}
\end{document}